\documentclass{article}

\usepackage{PRIMEarxiv}

\usepackage[utf8]{inputenc} % allow utf-8 input
\usepackage[T1]{fontenc}    % use 8-bit T1 fonts
\usepackage{hyperref}       % hyperlinks
\usepackage{url}            % simple URL typesetting
\usepackage{booktabs}       % professional-quality tables
\usepackage{amsfonts}       % blackboard math symbols
\usepackage{nicefrac}       % compact symbols for 1/2, etc.
\usepackage{microtype}      % microtypography
\usepackage{lipsum}
\usepackage{fancyhdr}       % header
\usepackage{graphicx}       % graphics
\graphicspath{{media/}}     % organize your images and other figures under media/ folder

\title{MO-CTranS: A unified multi-organ segmentation model learning from multiple heterogeneously labelled datasets}

\author{
Zhendi Gong, Andrew P. French, Xin Chen \\
  School of Computer Science \\
  University of Nottingham \\
  UK\\
  \texttt{\{Zhendi Gong\}psxzg6@nottingham.ac.uk} \\
   \And
  Susan Francis, Eleanor Cox \\
  School of Physics \\
  University of Nottingham \\
  UK\\
   \And
       \quad\quad Stamatios N. Sotiropoulos, Dorothee P. Auer \\
  \quad\quad School of Medicine \\
\quad\quad University of Nottingham \\
  \quad\quad UK \\    
   \And
Guoping Qiu \\
School of Computer Science \\
University of Nottingham Ningbo China \\
China
}

\begin{document}
\maketitle

\begin{abstract}
Multi-organ segmentation holds paramount significance in many clinical tasks. In practice, compared to large fully annotated datasets, multiple small datasets are often more accessible and organs are not labelled consistently. Normally, an individual model is trained for each of these datasets, which is not an effective way of using data for model learning. It remains challenging to train a single model that can robustly learn from several partially labelled datasets due to label conflict and data imbalance problems. We propose MO-CTranS: a single model that can overcome such problems. MO-CTranS contains a CNN-based encoder and a Transformer-based decoder, which are connected in a multi-resolution manner. Task-specific tokens are introduced in the decoder to help differentiate label discrepancies. Our method was evaluated and compared to several baseline models and state-of-the-art (SOTA) solutions on abdominal MRI datasets that were acquired in different views (i.e. axial and coronal) and annotated for different organs (i.e. liver, kidney, spleen). Our method achieved better performance (most were statistically significant) than the compared methods. Github link: \url{https://github.com/naisops/MO-CTranS}.
\end{abstract}

% keywords can be removed

\section{Introduction}
Multi-organ segmentation is a crucial yet challenging task, which is essential for a variety of clinical applications. With the success of convolutional neural networks (CNN), deep learning-based methods have achieved state-of-the-art (SOTA) performance in medical image segmentation \cite{gibson2018automatic,isensee2021nnu,wang2019abdominal}. Such methods typically require a large dataset that all organs of interest are annotated consistently. However, due to expertise and time constraints, it is extremely hard to obtain such a large-scale medical dataset. Instead, several small datasets are available from different sources with different organs of interest using different annotation strategies. Intuitively, training individual models on these datasets separately and then combining the predictions can solve the problem. However, it is time-consuming and computationally costly. Additionally, different datasets from the same imaging modality share some common information, which can not be fully utilised by training separate models. Thus, it is highly desirable to train a unified model that can learn from these multiple datasets simultaneously. One of the main challenges is label conflict. For instance, the labelled organ in one dataset can be considered as background in another, which would confuse the model in the learning process. Another potential issue is data imbalance across different datasets, which could cause performance inconsistency among organs. 

There are existing solutions to address the label conflict issue. We categorise them into two groups. The first group of methods focused on network design to build unified models. Med3D \cite{chen2019med3d} adopted a multi-head network, where the main component shares model weights learning across different datasets, and individual task-specific layers are separately trained on different datasets. However, such a method suffers from high computational costs and its scalability is poor when more datasets are included. Moreover, conditioned networks are commonly used for heterogeneously labelled image segmentation. Dmitriev et al. \cite{dmitriev2019learning} used conditions from label classes between convolutional layers to integrate with the intermediate activation signal for specific organ prediction. Similarly, Zhang et al. \cite{zhang2021multiorgan} proposed conditional nn-Unet, where they leveraged the conditional information of organ classes in each resolution level of the decoder for specific organ segmentation. Based on DoDNet \cite{zhang2021dodnet}, Xie et al. proposed TransDoDNet \cite{xie2023learning} by replacing the convolution-based kernel generator with a Transformer-based kernel generator. The generated kernels were used as weights in different convolutional layers for different segmentation tasks.

\begin{figure*}[!t]
\centering
\includegraphics[width=\textwidth]{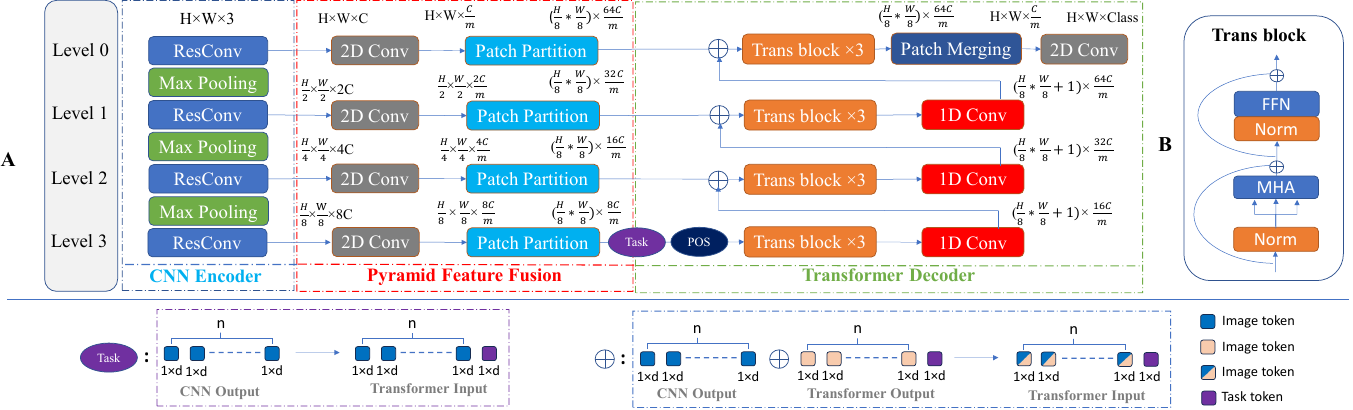}
\caption{\label{fig:network}Overview of MO-CTranS. * indicates value multiplication, and $x\times y$ means the feature dimension of $x$ ``by" $y$.}
\end{figure*}

The second group of methods utilised pseudo mask generation. Lian et al. \cite{lian2023learning} used partial labels before training a single-organ model and generating pseudo labels with mutual prior, then trained a multi-organ model based on the pseudo masks. Similarly, Huang et al. \cite{huang2020multi} first trained single-organ models to generate pseudo masks for unlabelled organs, then used co-training to train weight-averaged models based on the fully pseudo-annotated dataset. Such methods require multiple models to be trained, which are computationally expensive. To solve such a problem, Liu et al. \cite{liu2024cosst} proposed COSST, they introduced different supervision types to maximise the overlap between the prediction and its ground truth, minimising the overlap with the ground truth of different organs, and maximise the overlap between the prediction and its corresponding pseudo label. 

In this work, based on our previously developed CNN-Transformer segmentation architecture \cite{gong2024ctrans}, we add a task-specific token in the Transformer block to effectively differentiate different organ segmentation tasks. We evaluated our method and compared to SOTA solutions on 4 local partially labelled abdominal MRI datasets. The results demonstrate that our method achieves the best performance in most cases with statistical significance. The main contribution of our work is that we propose a novel and light model architecture that utilises a task token to learn from several datasets that are heterogeneously labelled. Our method does not require task-specific layers, separate decoder heads, or pseudo mask generation, which is simpler than many other SOTA methods. 

\section{Method}
\subsection{Problem definition and model overview}
It is assumed that multiple datasets are obtained through different sources which are heterogeneously labelled, and their acquisition view (i.e. axial, coronal or sagittal) could also be different. The $k^{th}$ dataset is denoted by $D_k=\{(x_i^k, y^k_i)\}$, where $x^k_i$ is the $i^{th}$ image sample, and $y^k_i$ is the corresponding label taking a value in $\mathcal{Y}_k \subset \mathcal{Y}$, where $\mathcal{Y}$ is the total label set (the same organ with different views has different labels). The goal is to train a unified model $f(x^k_i,\theta)$ using all $D_k$. In model inference, given an input image $x$, the model should be able to predict all labels in $\mathcal{Y}^{(v)}$, where $v$ indicates the view of the current input.

As depicted in Fig. \ref{fig:network} (A), MO-CTranS contains three main components, including a CNN-based encoder (CE), a pyramid feature fusion module (PFF), and a Transformer-based decoder (TD). CE is responsible for extracting image features from the input images in different scales. The PFF downsamples and reshapes the extracted features to the input image tokens of TD, and TD learns to capture the long-range relationships of these image tokens. We use the average of soft dice loss \cite{milletari2016v} and cross-entropy \cite{bengio2017deep} as the loss function for model training. 

\subsection{CNN-based encoder}
The encoder follows a similar network design of Residual U-Net \cite{zhang2018road}. Each ResConv block contains two convolutional layers, and a skip connection containing a convolutional layer is added between the input and output of each ResConv block. A batch normalisation layer and a ReLU activation layer are added after each convolutional layer. Furthermore, except for the last ResConv in level 3, a max pooling layer with the kernel size of $2\times2$ is used after the ResConv blocks to downsample the image features.

\subsection{Pyramid feature fusion}
To reduce the computational cost, PFF accepts the CNN-extracted image features and downsamples them along the channel dimension in a fixed ratio of $m$. In different levels, the Patch Partition operation divides the feature map into a fixed number of patches (a.k.a. tokens), which is determined in the deepest level, where each pixel in the feature map is considered as an image token. As shown in the example in Fig. \ref{fig:network} (A), the feature map in the size of $(H/8)\times(W/8)\times(8C/m)$ is divided into $(H/8)*(W/8)$ tokens with the size $1\times(8C/m)$, where $H$ and $W$ are the height and width of the input image.

\subsection{Transformer-based decoder}
As shown in Fig. \ref{fig:network} (A), each level in TD contains three Transformer blocks. Each Transformer block contains a multi-head self-attention block (MHA) and a feed forward network (FFN) as shown in Fig. \ref{fig:network} (B). The input tokens of the Transformer block are linearly projected to query (Q), key (K) and value (V). Next, MHA takes Q, K, and V as input, then outputs the weighted sum of V based on the attention score calculated from Q and K \cite{vaswani2017attention}. The output is then updated by FFN that contains several fully connected linear layers. Except for the deepest level, the input tokens in level $i$ are the sum of the output from PFF in level $i$ and the output from the Transformer blocks in level $i+1$. The 1D Conv in TD linearly projects the Transformer output to the same size of the PFF output in the shallower level. 

In the deepest level, the output image tokens in size $n \times d$ from PFF are inserted with a 1D task-specific token ($1 \times d$) as shown in Fig. \ref{fig:network}, where $n$ is the number of image tokens and $d$ is the token size. Then the $n+1$ tokens are added with the positional embedding in size $(n+1)\times d$ before input to the Transformer blocks. The input task token is predefined by the index of task (e.g, filled with 1 for the task 1) and fixed during model training. By inserting the task token, the image tokens and the task token can interact with each other by the self-attention layers in Transformer blocks. The image tokens from PFF are only added to the image tokens from Transformer blocks, and leave the task token unchanged, as shown in Fig. \ref{fig:network}. Thus, the output image tokens can interact with the image tokens from CNN in each resolution to reduce information loss. At last, only the image tokens are reshaped back using Patch Merging (reversed operation of Patch Partition) and processed by 2D Conv for image segmentation. Thus, the model can handle different tasks by inputting different task token as the prompt.

\section{Method Evaluation}
\subsection{Datasets}
For method evaluation, we constructed our own dataset that were collected from Sir Peter Mansfield Imaging Centre at University of Nottingham, UK. We will make the dataset publicly available for research purpose. 

The datasets (denoted as D1-D4) are multi-slice balanced turbo field echo (bTFE) localiser MRIs, each with approximately 35 slices of $1.75\times1.75\times7$ $mm^3$ resolution. D1 (18 subjects) was acquired on axial view for liver segmentation. D2 (17 subjects) was acquired on axial view for both liver and spleen segmentation. D3 (20 subjects) was acquired on coronal view for spleen segmentation and D4 (100 subjects) was acquired on coronal view for kidney segmentation. Label conflict appears in the datasets that are acquired in the same view. Additionally, we explicitly included much more examples in D4 (100) compared to D3 (20) in the coronal view. It helps to investigate if label conflict becomes a more severe problem when training using imbalanced datasets. Note that, we also had the annotation ground truth for spleen in D1, but only used it for evaluation not training. This is to evaluate the performance of segmenting unlabelled organs. 

\begin{table}[!t]
\caption{\label{tab1} Results of method comparison. \textbf{Bold} indicates the best performer, and $\dag$ indicates statistical significance compared to the bold results (p$<$0.05 measured by WSRT). The value in brackets for ASSD is the number of empty predictions. $Bs$, $Bm$, $Dm$, and $Dd$ refer to $Base_{single}$, $Base_{multi}$, $Dec_{multi}$, and $Dec_{dynam}$ respectively.}
\tiny
\resizebox{\columnwidth}{!}{%
\begin{tabular}{c|cc|cc|c|c}
\hline
 & \multicolumn{2}{c|}{D1 (Axial)} & \multicolumn{2}{c|}{D2 (Axial)} & \multicolumn{1}{c|}{D3 (Coronal)} & \multicolumn{1}{c}{D4 (Coronal)} \\
\cline{2-7}
 & \multicolumn{1}{c}{liver} & \multicolumn{1}{c|}{spleen} & \multicolumn{1}{c}{liver} & \multicolumn{1}{c|}{spleen} & \multicolumn{1}{c|}{spleen} & \multicolumn{1}{c}{kidney} \\
\cline{2-7}
& \multicolumn{6}{c}{DC} \\
\hline
Bs  & $\dag$0.832 & /        & $\dag$0.885 & $\dag$0.831 & $\dag$0.757 & $\dag$0.877 \\
Bm  & $\dag$0.855 & $\dag$0.374 & $\dag$0.909 & $\dag$0.359 & $\dag$0.257 & $\dag$0.768 \\
Dm  & $\dag$0.868 & $\dag$0.882 & $\dag$0.921 & $\dag$0.887 & $\dag$0.799 & $\dag$0.883 \\
Dd  & 0.875      & $\dag$0.890 & 0.923      & 0.891      & $\dag$0.805 & $\dag$0.876 \\
Ours & \textbf{0.875} & \textbf{0.903} & \textbf{0.923} & \textbf{0.894} & \textbf{0.817} & \textbf{0.883} \\
\hline
& \multicolumn{6}{|c}{ASSD} \\
\cline{1-7}
Bs  & $\dag$3.244(26) & /            & $\dag$3.598(6)  & $\dag$3.468(6)  & $\dag$7.884(0)  & $\dag$2.979(13) \\
Bm  & $\dag$2.782(20) & $\dag$4.195(131) & $\dag$2.378(3)  & $\dag$3.844(149) & $\dag$10.168(152) & $\dag$5.671(69) \\
Dm  & 2.597(15)      & $\dag$3.201(0)   & 1.970(3)       & $\dag$1.902(0)   & $\dag$4.604(6)   & 3.128(6) \\
Dd  & $\dag$2.815(12) & $\dag$3.817(0)   & $\dag$2.513(0)  & $\dag$2.172(1)   & $\dag$5.415(0)   & $\dag$3.270(2) \\
Ours & \textbf{2.255(19)} & \textbf{2.845(0)} & \textbf{1.809(3)} & \textbf{1.862(0)} & \textbf{3.205(10)} & \textbf{2.966(7)} \\
\hline
& \multicolumn{6}{|c}{Complexity \& training efficiency} \\
\cline{2-7}
 & \multicolumn{2}{c|}{NP} & \multicolumn{2}{c|}{GM} & \multicolumn{2}{c}{TS} \\
\cline{1-7}
Bs  & \multicolumn{2}{c|}{40.64} & \multicolumn{2}{c|}{4.27} & \multicolumn{2}{c}{0.30} \\
Bm  & \multicolumn{2}{c|}{40.64} & \multicolumn{2}{c|}{4.29} & \multicolumn{2}{c}{0.32} \\
Dm  & \multicolumn{2}{c|}{143.17} & \multicolumn{2}{c|}{5.63} & \multicolumn{2}{c}{0.90} \\
Dd  & \multicolumn{2}{c|}{40.86} & \multicolumn{2}{c|}{4.28} & \multicolumn{2}{c}{0.25} \\
Ours & \multicolumn{2}{c|}{40.64} & \multicolumn{2}{c|}{4.40} & \multicolumn{2}{c}{0.30} \\
\hline
\end{tabular}%
}
\end{table}

\subsection{Experimental design and parameter setting}
 We used MO-CTranS without inserting the task token as the baseline model (\textbf{Base}). For a fair comparison, instead of comparing with the SOTA models directly, we compared the performance of different SOTA solutions applied to \textbf{Base}. There are 4 segmentation tasks (denoted as $\mathcal{T}$) in our experiments, which are axial liver, axial spleen, coronal spleen, and coronal kidney segmentation respectively. The compared methods were implemented as below:
\begin{enumerate}
  \item $Base_{single}$: 3 binary-class \textbf{Base} (D1, D3, and D4) and 1 multi-class \textbf{Base} (D2) were trained separately on 4 datasets.
  \item $Base_{multi}$: A single \textbf{Base} was trained on the mix of 4 datasets. To reduce the impact of label conflict, we consistently used 5-class labels, including 4 foreground classes for $\mathcal{T}$ and one background class.
  \item $Dec_{multi}$: According to Med3D \cite{chen2019med3d}, a single \textbf{Base} with different decoders for different tasks was trained on the mix of 4 datasets.
  \item $Dec_{dynam}$: According to TransDoDNet \cite{xie2023learning}, we integrated a dynamic head into \textbf{Base}. First, the 2D Conv layer in the decoder in \textbf{Base} was replaced by dynamic convolutional layers. Additionally, a learnable task-specific embedding ($4\times d$) was input to a task-image cross-attention module to be updated with the output image tokens ($n\times d$) from the Transformer block in the deepest level of \textbf{Base}. Then the task embedding was further updated by a FFN, and each task token contained the weights and biases for the dynamic segmentation head for each task. A single $Dec_{dynam}$ was trained on the mix of 4 datasets.
\end{enumerate}

Note that, during the model training of our method, $Dec_{multi}$, and $Dec_{dyna}$, to ensure that each data is assigned with a single task in $\mathcal{T}$, we divided the multi-calss data (D2) to several binary-class data. Furthermore, when the model is trained on the mix of datasets, we assigned the subjects from the larger dataset (D4) a smaller weight when calculating the loss to reduce the data imbalance impact. All models were 2D-based, and we input the model with three continuous slices and then output the prediction of the middle slice in a sliding window manner. All images were reshaped to $256\times256$ and only slices with non-empty labels were used for training and evaluation. We used 5-fold cross validation for model comparison. In evaluation, all predictions were resized back to the original image size, and the mean Dice coefficient (DC) and average symmetric surface distance (ASSD) were reported. Specially, ASSD is only calculated on the non-empty predictions, the number of empty predictions is shown in Table \ref{tab1}. Wilcoxon Signed Rank Test (WSRT) was used to assess the statistical significance between the results of two methods. The number of learnable parameters (NP (million)), GPU memory consumption (GM (GB)), and training time per step (TS (seconds)) are reported to demonstrate the model's complexity and training efficiency.

The following parameters of all models were the same: $C_{base}=64$, $m=4$ and the number of levels was 4. All models were trained on a 12GB GeForce GTX 1080 Ti GPU with Adam optimiser with a learning rate of 0.0001 for 100 epochs. The model with the best performance on the validation set was saved for testing.

\subsection{Results}
The results of the 5-fold cross validation of different methods are shown in Table \ref{tab1}. Firstly, our method consistently outperforms the baseline solutions $Base_{single}$ and $Base_{multi}$ in all cases with statistical significance. $Base_{multi}$ can not effectively handle the data imbalance problem (i.e. Dice=0.257 in D3), and it mainly focuses on predicting the main class (i.e. kidney in D4). It shows that our method can solve the label conflict issue (i.e. spleen in D1 and D2) that impacts the $Base_{Multi}$ model significantly. Secondly, compared to the SOTA solutions $Dec_{multi}$ and $Dec_{dynam}$, our method performs significantly better on most cases. Specially, the performance difference between our method and the other solutions differs larger on spleen of D1 and D3, which demonstrates that our model has a superior capacity to leverage information from different datasets for predicting on unlabelled classes (spleen in D1) and relatively small datasets (spleen in D3). Furthermore, the complexity and training efficiency shown in Table \ref{tab1} shows that our method and $Dec_{dynam}$ have significantly lighter network than $Dec_{multi}$, and are in almost the same complexity with a single \textbf{Base}.

\begin{figure*}[!b]
\centering
\includegraphics[width=\textwidth]{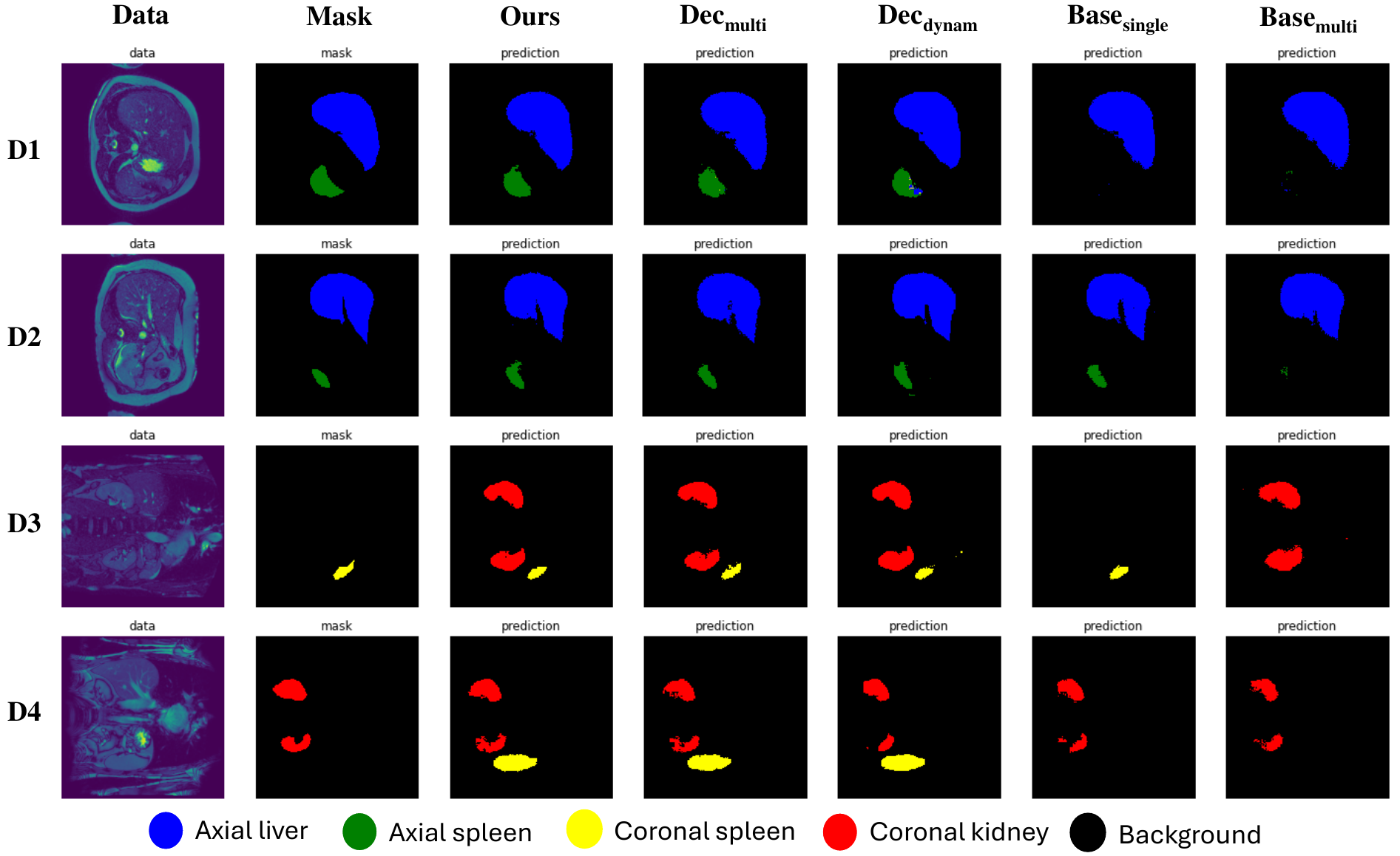}
\caption{\label{fig:data} Qualitative examples of different methods.}
\end{figure*}

Fig. \ref{fig:data} shows some qualitative examples to demonstrate the advantages of our method compared to the baseline models and the SOTA solutions. It can be seen that our model can predict all organs learnt from different datasets of the same view. This is not achievable by any of the baseline models. The image of D3 in the $Base_{Multi}$ column shows that the $Base_{Multi}$ model suffers from label conflict issues, especially if the datasets are imbalanced, i.e. the model predicted kidney instead of spleen in D3. Both SOTA solutions ($Dec_{multi}$ and $Dec_{dynam}$) can predict the unlabelled organs, but the overall performance is worse than ours.

\section{Conclusions}
This paper presents a novel CNN-Transformer model with the assistance of task tokens to achieve unified training based on several heterogeneously labelled datasets. It overcomes the issues of label conflict. This ability to learn jointly from several datasets helps to improve the performance of the baseline models that are trained individually for each dataset. Compared to the SOTA solutions, it achieves the best performance in most cases with a lighter network. In the future, we will include more datasets and improve the method to achieve continuous learning. 

\section*{Compliance with ethical standards}
Ethics approval was granted for the conduct of this human research (University of Nottingham, UK). All data was anonymized, and the participants’ information cannot be identified from the imaging data.

\section*{Acknowledgments}
This project is partly funded by UK Medical Research Council (Ref: MR/W014491/1).

%Bibliography
\bibliographystyle{unsrt}  
\bibliography{references}

\end{document}